\documentclass[11pt, a4paper]{article}
\usepackage{lrec2006}
\usepackage{graphicx}
\usepackage{multirow}

\title{Joint Named Entity Recognition and Stance Detection in Tweets}

\name{Dilek K\"u\c{c}\"uk}

\address{Electrical Power Technologies Group\\
T\"UB\.ITAK Energy Institute\\
Ankara, Turkey\\
dilek.kucuk@tubitak.gov.tr\\}

\abstract{
\vspace{1pt}
Named entity recognition (NER) is a well-established task of information extraction which has been studied for decades. More recently, studies reporting NER experiments on social media texts have emerged. On the other hand, stance detection is a considerably new research topic usually considered within the scope of sentiment analysis. Stance detection studies are mostly applied to texts of online debates where the stance of the text owner for a particular target, either explicitly or implicitly mentioned in text, is explored. In this study, we investigate the possible contribution of named entities to the stance detection task in tweets. We report the evaluation results of NER experiments as well as that of the subsequent stance detection experiments using named entities, on a publicly-available stance-annotated data set of tweets. Our results indicate that named entities obtained with a high-performance NER system can contribute to stance detection performance on tweets.
\newline
\newline
}

\begin{document}

\maketitleabstract

\section{Introduction}\label{intro}

With the emergence of social media applications like Twitter and Facebook, a considerable body of social media texts has accumulated. The need for the utilization of this user-generated content, for several purposes like trend analysis, has accelerated research on social media analysis.\\

As acknowledged in the related literature, the language used in social media texts is usually different from well-formed texts such as news articles, where the latter has been one of the most common text genre targeted by natural language processing (NLP) research so far. As a result, approaches proposed for well-formed texts have suffered from the \emph{porting problem}, revealed with poor performance on social media texts. One of these NLP problems is named entity recognition (NER) which targets at the extraction and classification of named entities like person, location, and organization names in texts \cite{Nadeau2007}. Several recent studies on NER report performance results on social media texts and propose customized systems for this text genre \cite{ritter2011named}. Another important research topic regarding social media analysis is sentiment analysis \cite{pang2008opinion} where the opinion or sentiment of the text owner is explored.\\

Stance detection is a considerably recent research field usually considered as a subproblem of sentiment analysis \cite{mohammad2016semeval}. In stance detection, the aim is to determine the stance of the text owner (as \emph{Favor}, \emph{Against}, or \emph{Neither}) for a particular target either explicitly or implicitly mentioned in the text \cite{mohammad2016semeval}. The most common text genres used by stance detection studies are on-line debates and social media texts like tweets. Some of the recent studies on this topic report performance evaluation results of different classifiers using different feature sets \cite{mohammad2016semeval} while others present publicly-available stance-annotated data sets \cite{mohammad2016dataset,sobhani2017dataset,kucuk2017stance}.\\

In this study, we present our experiments of using named entities for the purposes of improved stance detection in tweets. We have used the publicly-available tweet data set in Turkish annotated with stance information, together with the results of the corresponding SVM classifiers using unigrams as features in \cite{kucuk2017stance} as the baselines. We first perform NER on this data set and next use the named entities as additional features during our SVM-based stance detection experiments. Our findings are particularly encouraging as they provide evidence for the contribution of a high-performance NER system to the subsequent stance detection procedure using the extracted named entities. The rest of the paper is organized as follows: In Section 2, the evaluation results of an existing NER system on the data set are presented, the subsequent stance detection experiments using the named entities from the data set are described in Section 3, and finally Section 4 summarizes the paper together with future research directions.

\section{Named Entity Recognition in Tweets}\label{ner}

NER is an information extraction task that has been studied for decades, especially on well-formed texts like news articles. More recently, considerable number of studies on NER target at social media texts, like tweets. Yet, as emphasized in the related literature \cite{ritter2011named}, porting existing NER systems to social media texts results in poor performance. Therefore, related studies usually propose customized systems and\slash or annotated data sets (to be used during the training of supervised systems) for this new text genre.\\

In this study, we have used the stance-annotated tweet data set in Turkish \cite{kucuk2017stance} which includes 700 random tweets related to two sports clubs and these clubs constitute the stance targets. The data set is a balanced one in the sense that 175 tweets are in favor of \emph{Target-1} and 175 tweets are against \emph{Target-1}, while 175 tweets are in favor of and 175 are against \emph{Target-2}. There are no tweet instances annotated with the class \emph{Neither} in the data set \cite{kucuk2017stance}. The names of the two target clubs are explicitly mentioned in all of the tweets although some of these mentions are neologisms or contracted forms while some others have writing errors, as expected due to the peculiarities of language use in Twitter \cite{Kucuk2014_2}.\\

In order to create the NER answer key for this data set, we have annotated it with person, location, and organization names. The resulting named entity statistics are provided in Table \ref{tab:ner_stats}.

\begin{table}[h!]
\centering
\caption{Statistical Information About the Named Entities in the Tweet Data Set.}
\label{tab:ner_stats}
\begin{tabular}{|l|l|l|l|l|l|l|}
\hline
&&&\multicolumn{4}{|c|}{\textbf{\emph{\# of Named Entities}}} \\
\cline{4-7}
\emph{\textbf{Target}} & \emph{\textbf{Stance Class}} & \emph{\textbf{\# of Tweets}} & \emph{\textbf{Person}} & \emph{\textbf{Location}} & \emph{\textbf{Organization}} & \emph{\textbf{TOTAL}} \\
\hline
\multirow{2}{*}{Target-1} & Favor & 175 & 12 & 17 & 207 & 236 \\
\cline{2-7}
                            & Against & 175 & 70 & 4 & 221 & 295 \\
\hline
\multirow{2}{*}{Target-2} & Favor & 175 & 8 & 24 & 247 & 279 \\
\cline{2-7}
                            & Against & 175 & 69 & 18 & 277 & 364 \\
\hline
\multicolumn{2}{|l|}{\emph{\textbf{TOTAL}}} & 700 & 159 & 63 & 952 & 1,174\\
\hline
\end{tabular}
\end{table}

As the NER tool, we have used the extended version of the rule-based tool \cite{Kucuk2009} proposed for news articles in order to perform better on tweets, as presented in \cite{Kucuk2014_2}. These extensions include relaxing the capitalization constraint to extract entities all in lowercase as well and diacritics-based extension of the lexical resources of the tool since characters with diacritics such as \emph{\c{c}, {\i}, {\"o}, \c{s}} are commonly replaced with the corresponding characters such as \emph{c, i, o, s} in tweets. The evaluation results of the NER tool on the tweet data set are presented in Table \ref{tab:ner_results}, in terms of the corresponding metrics of precision (P), recall (R), and F-Measure (F), without giving credit to partial extractions. That is, a named entity extraction is considered as correct if both its type and all of its tokens are correctly identified by the system.

\begin{table}[h!]
\centering
\caption{Evaluation Results of the NER Tool on the Tweet Data Set.}
\label{tab:ner_results}
\begin{tabular}{|l|l|p{0.65in}|p{0.65in}|p{0.65in}|}
\hline
\emph{\textbf{Target}} & \emph{\textbf{Stance Class}} & \emph{\textbf{P (\%)}} & \emph{\textbf{R (\%)}} & \emph{\textbf{F (\%)}} \\
\hline
\multirow{2}{*}{Target-1} & Favor & 73.79 & 64.41 & 68.78 \\
\cline{2-5}
                            & Against & 77.14 & 36.61 & 49.66 \\
                             \hline
\multirow{2}{*}{Target-2} & Favor & 78.80 & 51.97 & 62.63 \\
\cline{2-5}
                            & Against & 71.01 & 40.38 & 51.49 \\
\hline
\end{tabular}
\end{table}

The NER results obtained are not favorable compared to the NER results on news articles in Turkish (usually over 85\% in F-Measure) but as reported in the literature, NER performance on tweets is usually considerably lower due to the particular language use in tweets. For instance, using the same version of the NER tool, the best F-Measure rate obtained on another tweet data set in Turkish is 48.13\% \cite{Kucuk2014_2}. The F-Measure rates given in Table \ref{tab:ner_results} are all higher than 48.13\%, reaching to 68.78\% for the tweets marked as in favor of \emph{Target-1}, and therefore these rates can be considered encouraging. Yet, we should also note a difference between the features of the tweet data sets in \cite{Kucuk2014_2} and in the current study. As pointed out at the beginning of this section, in each of the tweets of the stance data set used in the current study, the sports clubs (a named entity of organization type) which are the stance targets are explicitly mentioned while there is no such restriction for the data sets used in \cite{Kucuk2014_2}. Hence, the data set used in the current study can better be considered as a ``targeted tweet data set" and this may be one of the reasons for the high rates compared to those reported in \cite{Kucuk2014_2}.\\

The results in Table \ref{tab:ner_results} also indicate that NER performance is higher for the tweets marked as in favor of their targets when compared with the tweets marked as against. The recall rates are particularly low for this latter set of tweets and one of the reasons for this observation may be the users' common employment of neologisms (in the negative sense) especially for the tweet targets, which are missed by the NER tool.\\

To summarize, we have annotated the stance data set with named entities in order to create the NER answer key and performed NER evaluations on this data set. The results obtained are promising and even higher than those results reported for similar settings in the literature. The evaluation results obtained in the current study are significant as a new set of NER evaluations on a proprietary ``targeted" tweet data set in Turkish, in addition to the data sets used in studies such as \cite{Kucuk2014_2}, since NER on tweets is still an important research field of NLP.

\section{Using Named Entities for Stance Detection in Tweets}\label{sd}

In the current study, we have used the stance-annotated tweet data set described in \cite{kucuk2017stance}. Also presented in \cite{kucuk2017stance} are the results of the following experiments on this data set:
\begin{itemize}
    \item   SVM classifiers using unigrams as features,
    \item   SVM classifiers using bigrams as features,
    \item   SVM classifiers using unigrams and the existence of hashtags in tweets as features.
\end{itemize}
The corresponding results have indicated that using unigrams as features leads to favorable performance rates and using unigrams together with hashtag features improves these results further, while using bigrams as features of the SVM classifiers results in poor performance \cite{kucuk2017stance}. The favorable results corresponding to the former two settings are provided in Table \ref{tab:sd_baseline} as excerpted from \cite{kucuk2017stance}, in order to be used as reference results for comparison purposes. These are 10-fold cross validation results on the data set. As can be observed in Table \ref{tab:sd_baseline}, using the existence of hashtags as an additional feature improves stance detection performance in terms of average F-Measure for \emph{Target-2} although it leads to a slight decrease in F-Measure for \emph{Target-1} \cite{kucuk2017stance}.

\begin{table}[h!]
\centering
\caption{Previous Evaluation Results of the SVM Classifiers for Stance Detection with Features Based on Unigrams and Unigrams$+$Hashtags (as Reported in \cite{kucuk2017stance}).}
\label{tab:sd_baseline}
\begin{tabular}{|l|l|p{0.65in}|p{0.65in}|p{0.65in}|p{0.65in}|p{0.65in}|p{0.65in}|}
\hline
&&\multicolumn{3}{|c|}{\emph{\textbf{Unigram-Based Features}}}&\multicolumn{3}{|c|}{\emph{\textbf{Unigram$+$Hashtag-Based Features}}}\\
\cline{3-8}
\emph{\textbf{Target}} & \emph{\textbf{Stance Class}} & \emph{\textbf{P (\%)}} & \emph{\textbf{R (\%)}} & \emph{\textbf{F (\%)}} & \emph{\textbf{P (\%)}} & \emph{\textbf{R (\%)}} & \emph{\textbf{F (\%)}} \\
\hline
\multirow{3}{*}{Target-1} & Favor & 75.2 & 92.0 & 82.8 & 75.0 & 90.9 & 82.2 \\
                            & Against & 89.7 & 69.7 & 78.5 & 88.4 & 69.7 & 78.0 \\
                             & \textbf{Average} & \textbf{82.5} & \textbf{80.9} & \textbf{80.6} & \textbf{81.7} & \textbf{80.3} & \textbf{80.1} \\
                             \hline
\multirow{3}{*}{Target-2} & Favor & 68.5 & 83.4 & 75.3 & 70.0 & 85.1 & 76.8 \\
                            & Against & 78.8 & 61.7 & 69.2 & 81.0 & 63.4 & 71.2 \\
                             & \textbf{Average} & \textbf{73.7} & \textbf{72.6} & \textbf{72.2} & \textbf{75.5} & \textbf{74.3} & \textbf{74.0} \\
\hline
\end{tabular}
\end{table}

In this study, we investigate the possible contribution of named entities to stance detection in tweets. We do not consider named entity types as features, but instead we use named entities as additional features for SVM classifiers which have used unigrams as features. Named entities which are not inflected and which comprise only single tokens are no different than existing unigrams. However, there are named entities in the form of ngrams. Additionally, Turkish is an agglutinative language and the particular NER tool that we have employed extracts named entities in their bare forms by excluding the sequence of suffixes attached to named entities, making the corresponding single-token named entities different from unigrams. We should note that during the named entity annotation procedure to create the answer key (as explained in Section 2), bare forms of the entities are annotated, making the system results and the annotations consistent.\\

Similar to the settings in \cite{kucuk2017stance}, we have used the SVM classifier based on the SMO algorithm \cite{platt1999}, available in the Weka tool \cite{weka2009}, during our stance detection experiments. 10-fold cross-validation results of the classifiers using the named entities extracted by the employed NER tool from the data set are provided in Table \ref{tab:sd_new_system} while the corresponding results of the classifiers using the named entities in the manually-annotated version of the data set (i.e., the answer key for the NER procedure) are provided in Table \ref{tab:sd_new_answer}.

\begin{table}[h!]
\centering
\caption{Evaluation Results of the SVM Classifiers for Stance Detection with Features Based on Unigrams$+$Named Entities and Unigrams$+$Named Entities$+$Hashtags, with Named Entities Extracted by the NER Tool.}
\label{tab:sd_new_system}
\begin{tabular}{|l|l|p{0.65in}|p{0.65in}|p{0.65in}|p{0.65in}|p{0.65in}|p{0.65in}|}
\hline
&&\multicolumn{3}{|c|}{\emph{\textbf{Unigrams$+$Named Entity-Based}}}&\multicolumn{3}{|c|}{\emph{\textbf{Unigrams$+$Named Entity}}}\\
&&\multicolumn{3}{|c|}{\emph{\textbf{Features}}}&\multicolumn{3}{|c|}{\emph{\textbf{$+$Hashtag-Based Features}}}\\
\cline{3-8}
\emph{\textbf{Target}} & \emph{\textbf{Stance Class}} & \emph{\textbf{P (\%)}} & \emph{\textbf{R (\%)}} & \emph{\textbf{F (\%)}} & \emph{\textbf{P (\%)}} & \emph{\textbf{R (\%)}} & \emph{\textbf{F (\%)}} \\
\hline

\multirow{3}{*}{Target-1} & Favor & 75.1 & 89.7 & 81.8 & 75.6 & 90.3 & 82.3 \\
                            & Against & 87.2 & 70.3 & 77.8 & 87.9 & 70.9 & 78.5 \\
                             & \textbf{Average} & \textbf{81.2} & \textbf{80.0} & \textbf{79.8} & \textbf{81.8} & \textbf{80.6} & \textbf{80.4} \\
                             \hline
\multirow{3}{*}{Target-2} & Favor & 72.2 & 87.4 & 79.1 & 71.8 & 84.6 & 77.7 \\
                            & Against & 84.1 & 66.3 & 74.1 & 81.3 & 66.9 & 73.4 \\
                             & \textbf{Average} & \textbf{78.1} & \textbf{76.9} & \textbf{76.6} & \textbf{76.5} & \textbf{75.7} & \textbf{75.5} \\
\hline
\end{tabular}
\end{table}

\begin{table}[h!]
\centering
\caption{Evaluation Results of the SVM Classifiers for Stance Detection with Features Based on Unigrams$+$Named Entities and Unigrams$+$Named Entities$+$Hashtags, with Named Entities in the Manually-Annotated Answer Key.}
\label{tab:sd_new_answer}
\begin{tabular}{|l|l|p{0.65in}|p{0.65in}|p{0.65in}|p{0.65in}|p{0.65in}|p{0.65in}|}
\hline
&&\multicolumn{3}{|c|}{\emph{\textbf{Unigrams$+$Named Entity-Based}}}&\multicolumn{3}{|c|}{\emph{\textbf{Unigrams$+$Named Entity}}}\\
&&\multicolumn{3}{|c|}{\emph{\textbf{Features}}}&\multicolumn{3}{|c|}{\emph{\textbf{$+$Hashtag-Based Features}}}\\
\cline{3-8}
\emph{\textbf{Target}} & \emph{\textbf{Stance Class}} & \emph{\textbf{P (\%)}} & \emph{\textbf{R (\%)}} & \emph{\textbf{F (\%)}} & \emph{\textbf{P (\%)}} & \emph{\textbf{R (\%)}} & \emph{\textbf{F (\%)}} \\
\hline

\multirow{3}{*}{Target-1} & Favor & 74.9 & 92.0 & 82.6 & 76.3 & 92.0 & 83.4 \\
                            & Against & 89.6 & 69.1 & 78.1 & 89.9 & 71.4 & 79.6 \\
                             & \textbf{Average} & \textbf{82.3} & \textbf{80.6} & \textbf{80.3} & \textbf{83.1} & \textbf{81.7} & \textbf{81.5} \\
                             \hline

\multirow{3}{*}{Target-2} & Favor & 74.4 & 91.4 & 82.1 & 74.3 & 90.9 & 81.7 \\
                            & Against & 88.9 & 68.6 & 77.4 & 88.2 & 68.6 & 77.2 \\
                             & \textbf{Average} & \textbf{81.7} & \textbf{80.0} & \textbf{79.7} & \textbf{81.3} & \textbf{79.7} & \textbf{79.5} \\
\hline
\end{tabular}
\end{table}

Based on the results presented in Tables \ref{tab:sd_baseline}, \ref{tab:sd_new_system}, and \ref{tab:sd_new_answer}, the following conclusions can be drawn:

\begin{itemize}
    \item   Using named entities as additional features improves the stance detection performance considerably for \emph{Target-2} although a slight performance decrease is observed for \emph{Target-1}, when compared with the case in which only the unigrams are used as features. These results indicate that stance detection task can benefit from the outputs of NER tools.
    \item   Using manually annotated named entities (in the answer key) improves stance detection performance for both targets and for both stance classes when compared with using named entities extracted with a NER tool. This is an expected result, since there are errors in the output of the NER tool (as quantified in Table \ref{tab:ner_results}) where some entities are missed and some other token sequences are incorrectly extracted as named entities. This finding provides evidence for the contribution of a high-performance NER tool to the stance detection task. The less errors the employed NER tool makes, the more successful the stance detection system, utilizing the output of the NER tool, will be.
    \item   Joint use of named entities and existence of hashtags as additional features to unigrams improves stance detection performance slightly for \emph{Target-1} while slight decreases are observed for \emph{Target-2} in this settings. Hence, further experiments are necessary to make sound conclusions regarding joint utilization of named entities and hastags as features of the SVM classifiers for the stance detection task.
    \item   As has been reported in \cite{kucuk2017stance}, the overall evaluation results of the stance detection task are considerably higher for the \emph{Favor} class when compared with the results for the \emph{Against} class, in all settings given in Table \ref{tab:sd_new_system} and \ref{tab:sd_new_answer}.
\end{itemize}

\section{Conclusion}\label{conc}
Stance detection is a relatively recent research topic similar in nature to sentiment analysis. The aim of stance detection is to determine the stance of a text owner for a particular target and stance detection modules can be used within several different social media analysis applications. In this study, we have first performed NER on a stance-annotated tweet data set in Turkish and then investigated the possible contribution of named entities as SVM features for the stance detection task. During this procedure, we have first reported the evaluation results of a NER tool that has been extended to be perform better on tweets and then have utilized the extracted named entities as additional features to SVM classifiers which already employ unigrams as features. Our results indicate that named entities can considerably improve the stance detection performance when used together with unigrams. Future work includes testing the same classifier settings on larger data sets, and performing similar tests by using stance-annotated data sets and NER tools for other languages such as English and compare the corresponding test results with the ones in the current study.

\bibliographystyle{lrec2006}

\begin{thebibliography}{}

\bibitem[\protect\citename{Hall \bgroup et al.\egroup }2009]{weka2009}
Mark Hall, Eibe Frank, Geoffrey Holmes, Bernhard Pfahringer, Peter Reutemann,
  and Ian~H Witten.
\newblock 2009.
\newblock The weka data mining software: An update.
\newblock {\em ACM SIGKDD explorations newsletter}, 11(1):10--18.

\bibitem[\protect\citename{K\"u\c{c}\"uk and Steinberger}2014]{Kucuk2014_2}
Dilek K\"u\c{c}\"uk and Ralf Steinberger.
\newblock 2014.
\newblock Experiments to improve named entity recognition on {Turkish} tweets.
\newblock In {\em Proceedings of the EACL Workshop on Language Analysis for
  Social Media}, pages 71--78.

\bibitem[\protect\citename{K\"u\c{c}\"uk and Yaz{\i}c{\i}}2009]{Kucuk2009}
Dilek K\"u\c{c}\"uk and Adnan Yaz{\i}c{\i}.
\newblock 2009.
\newblock Named entity recognition experiments on {Turkish} texts.
\newblock In {\em Proceedings of the International Conference on Flexible Query
  Answering Systems}, volume 5822 of {\em LNCS}, pages 524--535.

\bibitem[\protect\citename{K\"u\c{c}\"uk}2017]{kucuk2017stance}
Dilek K\"u\c{c}\"uk.
\newblock 2017.
\newblock Stance detection in {Turkish} tweets.
\newblock In {\em Proceedings of the International Workshop on Social Media
  World Sensors (SIDEWAYS) of the ACM Conference on Hypertext and Social Media
  (HT)}.

\bibitem[\protect\citename{Mohammad \bgroup et al.\egroup
  }2016a]{mohammad2016dataset}
Saif~M Mohammad, Svetlana Kiritchenko, Parinaz Sobhani, Xiaodan Zhu, and Colin
  Cherry.
\newblock 2016a.
\newblock A dataset for detecting stance in tweets.
\newblock In {\em Proceedings of the Language Resources and Evaluation
  Conference (LREC)}.

\bibitem[\protect\citename{Mohammad \bgroup et al.\egroup
  }2016b]{mohammad2016semeval}
Saif~M Mohammad, Svetlana Kiritchenko, Parinaz Sobhani, Xiaodan Zhu, and Colin
  Cherry.
\newblock 2016b.
\newblock Semeval-2016 task 6: Detecting stance in tweets.
\newblock In {\em Proceedings of the International Workshop on Semantic
  Evaluation, SemEval}.

\bibitem[\protect\citename{Nadeau and Sekine}2007]{Nadeau2007}
David Nadeau and Satoshi Sekine.
\newblock 2007.
\newblock A survey of named entity recognition and classification.
\newblock {\em Lingvisticae Investigationes}, 30(1):3--26.

\bibitem[\protect\citename{Pang and Lee}2008]{pang2008opinion}
Bo~Pang and Lillian Lee.
\newblock 2008.
\newblock Opinion mining and sentiment analysis.
\newblock {\em Foundations and Trends in Information Retrieval}, 2(1-2):1--135.

\bibitem[\protect\citename{Platt}1999]{platt1999}
John~C. Platt.
\newblock 1999.
\newblock Fast training of support vector machines using sequential minimal
  optimization.
\newblock {\em Advances in Kernel Methods}, pages 185--208.

\bibitem[\protect\citename{Ritter \bgroup et al.\egroup }2011]{ritter2011named}
Alan Ritter, Sam Clark, Oren Etzioni, et~al.
\newblock 2011.
\newblock Named entity recognition in tweets: an experimental study.
\newblock In {\em Proceedings of the Conference on Empirical Methods in Natural
  Language Processing}, pages 1524--1534.

\bibitem[\protect\citename{Sobhani \bgroup et al.\egroup
  }2017]{sobhani2017dataset}
Parinaz Sobhani, Diana Inkpen, and Xiaodan Zhu.
\newblock 2017.
\newblock A dataset for multi-target stance detection.
\newblock In {\em Proceedings of the 15th Conference of the European Chapter of
  the Association for Computational Linguistics: Volume 2, Short Papers}, pages
  551–--557.

\end{thebibliography}

\end{document}